\documentclass{article}    % Use this line for a4 paper
\usepackage[paper=a4paper,top=1in,bottom=1in,right=0.5in,left=1.5in]{geometry}
\usepackage{graphics} % for pdf, bitmapped graphics files
\usepackage{epsfig} % for postscript graphics files
\usepackage{mathptmx} % assumes new font selection scheme installed
\usepackage{times} % assumes new font selection scheme installed
\usepackage{amsmath} % assumes amsmath package installed
\usepackage{amssymb}  % assumes amsmath package installed
\usepackage{todonotes}

\usepackage{tikz,graphics,color,fullpage,float,epsf,subcaption, caption}
\usepackage{hyperref}
\usepackage[linesnumbered,ruled,vlined]{algorithm2e}
\usepackage{algorithmic}
\usetikzlibrary{bayesnet}
\usetikzlibrary{arrows}
\tikzset{>=latex}
\DeclareMathOperator{\E}{\mathbb{E}}

\usepackage{bm}
\usepackage{multirow}
\SetKwInput{KwInput}{Input}                % Set the Input
\SetKwInput{KwOutput}{Output}  
\usepackage{booktabs}
\usepackage{cuted}    % <===

\title{\LARGE \bf
PlaNet-ClothPick: Effective Fabric Flattening Based on Latent Dynamic Planning
}

\author{Halid Abdulrahim Kadi$^{1,*}$ and Kasim Terzi{\' c}$^{1}$% <-this % stops a space
\thanks{$^{1}$ School of Computer Science, University of St Andrews, Jack Cole Building, North Haugh, St Andrews, KY16 9SX United Kingdom}%
\thanks{$^{*}$ Correspondence author, E-mail: {\tt\small ah390@st-andrews.ac.uk;} Tel.: {\tt\small+44 1334 46 1630}}}

\begin{document}
\maketitle
\thispagestyle{empty}
\pagestyle{empty}

%%%%%%%%%%%%%%%%%%%%%%%%%%%%%%%%%%%%%%%%%%%%%%%%%%%%%%%%%%%%%%%%%%%%%%%%%%%%%%%%

\begin{abstract}
Why do Recurrent State Space Models such as PlaNet fail at cloth manipulation tasks? Recent work has attributed this to the blurry prediction of the observation, which makes it difficult to plan directly in the latent space. This paper explores the reasons behind this by applying PlaNet in the pick-and-place fabric-flattening domain. We find that the sharp discontinuity of the transition function on the contour of the fabric makes it difficult to learn an accurate latent dynamic model, causing the MPC planner to produce pick actions slightly outside of the article. By limiting picking space on the cloth mask and training on specially engineered trajectories, our mesh-free PlaNet-ClothPick surpasses visual planning and policy learning methods on principal metrics in simulation, achieving similar performance as state-of-the-art mesh-based planning approaches. Notably, our model exhibits a faster action inference and requires fewer transitional model parameters than the state-of-the-art robotic systems in this domain. Other supplementary materials are available at:
\href{https://sites.google.com/view/planet-clothpick}{https://sites.google.com/view/planet-clothpick}.
\end{abstract}

%%%%%%%%%%%%%%%%%%%%%%%%%%%%%%%%%%%%%%%%%%%%%%%%%%%%%%%%%%%%%%%%%%%%%%%%%%%%%%%%

\section{Introduction}

Deep reinforcement learning methods based on the Recurrent State Space Model
(RSSM), such as PlaNet \cite{hafner2019learning} and Dreamer \cite{hafner2020dream, hafner2021mastering, hafner2023mastering} have achieved state-of-the-art (SoTA) asymptotic performance and data efficiency in both continuous control \texttt{dm\_control} \cite{tassa2018deepmind} and discrete-action Atari 2600 \cite{bellemare2013arcade} benchmark environments. However, many authors have noted that RSSM-based models struggle with a canonical task in cloth-shaping: fabric flattening, where one or more end-effectors operate on a piece of square fabric to unfold it on a surface \cite{sun2013heuristic, seita2020deep, hoque2022visuospatial, kadi2023data}.

Most successful data-driven methods, such as imitation learning \cite{seita2020deep, seita2021learning} and reinforcement learning \cite{wu2019learning, hoque2022visuospatial, ma2022learning, lin2022learning, huang2022mesh} for fabric-flattening focus on quasi-static pick-and-place (P\&P) manipulation. Despite pick-and-fling and pick-and-blow primitives being operationally more effective than quasi-static P\&P primitives \cite{xu2022dextairity, ha2022flingbot}, P\&P is cost-effective as it only requires one robot arm and a camera.

Deep Planning Network (PlaNet) \cite{hafner2019learning} is a model-based reinforcement learning algorithm that uses model-predictive control (MPC) to plan on a latent dynamic model (LDM) trained based on the RSSM. However, PlaNet keeps failing on fabric flattening \cite{ma2022learning, yan2021learning, lin2020softgym}; it has been argued that this may be due to the blurry observation reconstruction of the LDM \cite{hoque2022visuospatial, lin2022learning}. In this paper, we investigate PlaNet's performance on the domain in simulation benchmark SoftGym \cite{lin2020softgym} to understand the causes of poor performance on this task. We note four contributions of this paper:

\begin{figure}[t]
    \centering
    \includegraphics[width=0.8\linewidth]{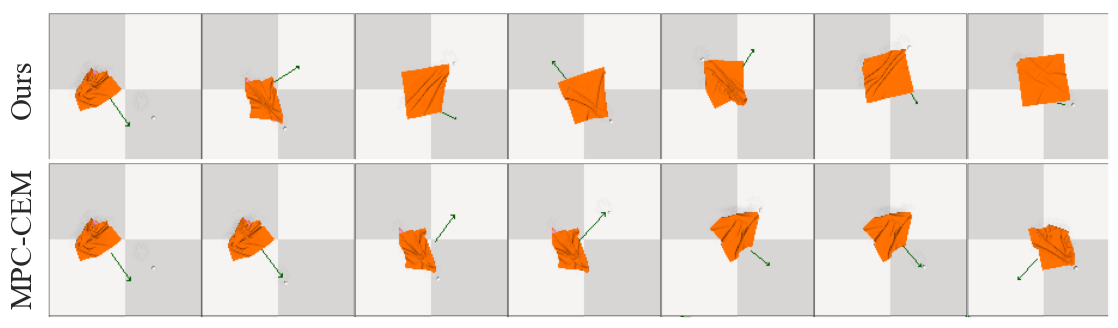}
    \caption{Flattening trajectories of ClothMaskPick-MPC (ours) and MPC with cross entropy method (MPC-CEM). The head and end of the green arrows represent each step's pick and place positions. The fundamental reason PlaNet fails in cloth flattening is that the latent dynamic model cannot accurately model the transition function's sharp discontinuity on the cloth's contour. By limiting the first picking sampled actions to fall inside the cloth mask, PlaNet-ClothPick achieves SoTA performance in this domain.}
    \label{fig:traj-compare}
\end{figure}

(1) We propose a new reward function for fabric flattening, which leads to better performance than the normalised coverage and the reward adopted by Hoque et al. (2022) \cite{hoque2022visuospatial}.

(2)  Inspired by Lin et al. (2022) \cite{lin2022learning} and Hoque et al. (2022) \cite{hoque2022visuospatial}, we suggest a domain-specific planning method ClothMaskPick-MPC that samples the first pick action on the cloth-mask to improve planning accuracy and efficiency (Figure \ref{fig:traj-compare}); this helps to overcome the difficulty in accurately modelling the sharp discontinuity on the article's contour.

(3) LDMs need a large amount of data to overcome the enormity and complexity of the cloth's dynamic. Apart from the expert, random, corner-biased trajectories \cite{hoque2022visuospatial}, we collect trajectories where the policy samples small dragging actions on an almost flattened fabric. This helps to boost the robustness and performance of the ClothMaskPick-MPC on the LDM. Besides, we employ data augmentation techniques, such as rotation \cite{lee2021learning} and flipping, to upsample the training trajectories to improve the robustness of the model.

(4) We also observe that PlaNet's LDM cannot learn a good latent prior distribution due to the complex non-linear behaviour of cloth-like deformable objects (CDOs), so we adopt \textit{KL balancing} \cite{hafner2021mastering} to improve both posterior and prior learning quality. We further improve the planning performance by incorporating prior reward learning.

We demonstrate that these four improvements lead to PlaNet-ClothPick achieving SoTA performance compared to mesh-based planning methods regarding primary metrics -- normalised coverage (NC) and normalised improvement (NI) against action steps, and it outperforms visual planning and policy learning methods (Figure \ref{fig:DRL-compare}). It also showcases a one-order-of-magnitude advantage regarding the action inference time and transitional model parameters compared to the previous SoTA robotic systems in this domain (Figure \ref{fig:SoTA-flattening-robotic}). The strong inductive biases of our method are introduced in ClothMaskPick-MPC and the specially engineered offline dataset, hence the name PlaNet-ClothPick. This paper shows that RSSM-based algorithms can play an important role in a wider range of application domains.

\section{Related Work}
%\emph{MPC in Pick-and-Place Cloth-flattening}
Model-based reinforcement learning (MBRL) applications in P\&P cloth-flattening are mainly Type I \cite{moerland2020framework}: planning or trajectory optimisation algorithms, where the agent needs access to the dynamic model of the environment for generating imaginary rollouts. The dynamic model can be either a known dynamic or a learned dynamic. Planning algorithms used in fabric-flattening are mainly based on model-predictive control (MPC), which can be further categorised into goal-conditioned MPC \cite{hoque2022visuospatial, yan2021learning} and reward-based MPC systems \cite{ma2022learning, lin2022learning, huang2022mesh}. 

Visual Foresight (VSF) \cite{hoque2022visuospatial} by Hoque et al. (2022) and the Contrastive Forward Model (CFM) \cite{yan2021learning} by Yan et al. (2020) are the two example applications of goal-conditioned MPC to fabric flattening, where the cost function is calculated based on the difference between the current and goal states. Note that VSF's cost function is calculated according to  Visual MPC \cite{ebert2018visual} framework, while the one of CFM is the distance of the two states at the latent space. 

reward-based MPC, on the other hand, selects top trajectories based on reward prediction from the prior rollout trajectories \cite{hafner2019learning, ma2022learning, lin2022learning}. In contrast to goal-conditioned MPC, the application domain of reward-based MPC is limited by the reward prediction function given to the algorithm. DefOrmable Object Manipulation (G-DOOM) is a latent reward-based MPC method that generates the prior rollout trajectories in the latent space trained with unsupervised-keypoint graph dynamics \cite{ma2022learning}. In contrast, Visible Connectivity Dynamics (VCD) by Lin et al. (2022) \cite{lin2022learning} is a mesh-based reward-based MPC method that applies rollout on the reconstructed mesh representation using a learned mesh dynamic \cite{pfaff2021learning}; they also proposed VCD Graph Imitation (VCD-GI), where a teacher dynamic model learns with the complete information of the cloth and distil the knowledge to the vision-based student. 

Mesh-based reward-based MPC methods, such as VCD, VCD-GI \cite{lin2022learning}, and MEDOR \cite{huang2022mesh}, outperform goal-conditioned MPC and latent reward-based MPC methods in cloth-flattening. Although they are invariant to the cloth shape, colour and camera pose, these methods cannot be easily applied to manipulating other kinds of objects, since the dynamic model is specially trained for CDOs. Most recent robotic systems focus on closing the simulation-to-reality gap of the mesh-based planning methods on garment-flattening tasks \cite{huang2023self, canberk2023cloth, wang2023trtm} by improving the mesh tracking accuracy in real-world trials.

Deep Planning Network (PlaNet) \cite{hafner2019learning} is a latent reward-based MPC method that employs a learned latent dynamic based on a Recurrent State Space Model (RSSM). While it performs well on continuous control benchmark environments like the \texttt{dm\_control} suite \cite{tassa2018deepmind}, numerous experiments have found it unsuitable for fabric flattening  \cite{yan2021learning, ma2022learning, lin2020softgym}. A possible reason is that the reconstructed observation from the visual model is fuzzy, which makes planning based on reconstructed vision hard due to the lack of precision around the edges and corners of the article \cite{lin2022learning}. 
\begin{figure*}[t]%
\centering
\subfloat[\centering Training of RSSM]{{\includegraphics[width=0.5\linewidth]{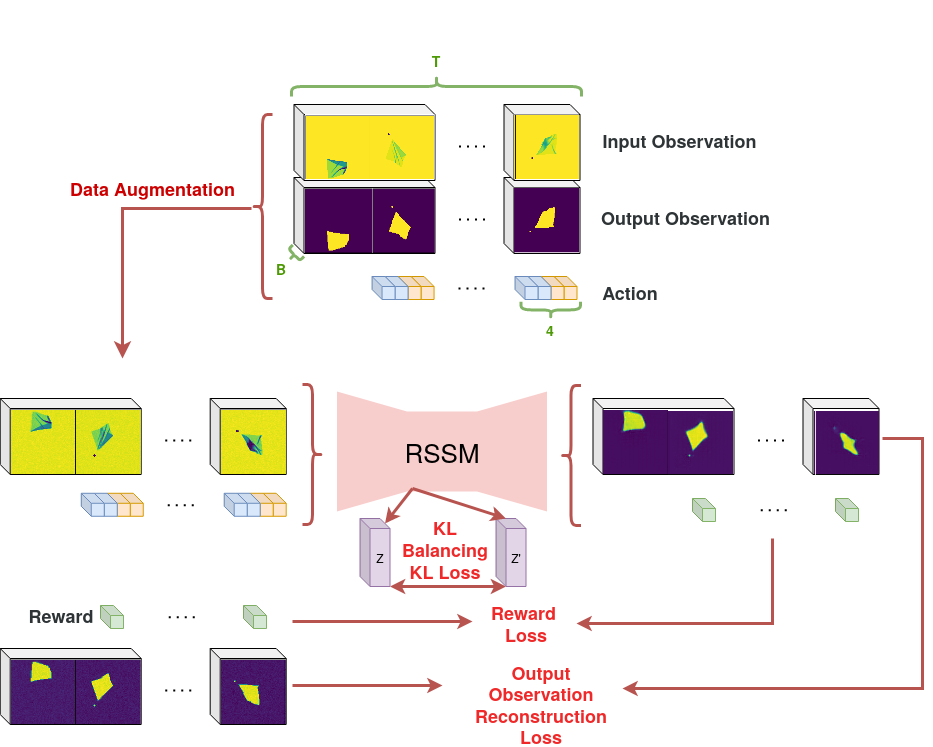}}}
\quad
\subfloat[\centering Action Inference of ClothMaskPick-MPC]{{\includegraphics[width=0.35\linewidth]{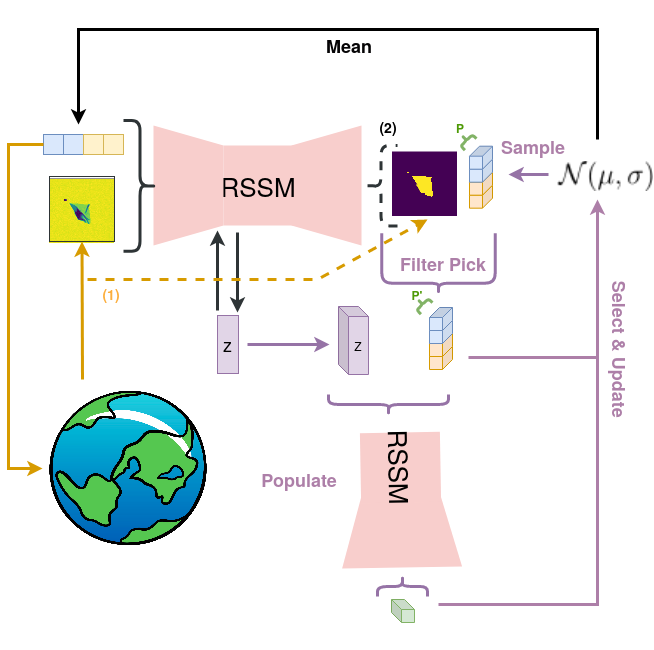}}}

    \caption{PlaNet-ClothPick. We use I2O to denote the different variants; for example, D2Mask represents the model's input as a depth image and the output as a cloth-mask image. The red line depicts training data flow, the purple line represents action optimization in planning, the black line signifies internal state updates, and the yellow line illustrates environment-agent input/output. In training, PlaNet-ClothPick applies batch-wise rotation and vertical flipping 
    on the input/output observations and actions before feeding them into the RSSM model; observation noise is only applied if the observations are RGB and/or depth images. During the inference time, ClothMaskPick-MPC
    samples pick-and-place actions from a normal distribution (initialised with mean as 0 and standard deviation as 1), then it filters them through an estimated cloth mask, which can be obtained in two different ways: (1) thresholding from the depth image of the environment or (2) predicting from the RSSM model if the decoding includes mask prediction. Then, it selects the top 10\% candidates based on the reward the RSSM predicted from the last-step posterior latent state and the sampled actions to update the normal distribution for the next optimisation iteration. After the planning, the method uses the mean of the distribution as an action to execute.}%
    \label{fig:method}%
\end{figure*}

Dreamer \cite{hafner2020dream, hafner2021mastering, hafner2023mastering} is a model-based actor-critic (AC) reinforcement learning algorithm that uses RSSM \cite{hafner2019learning} for representation learning. It shows significant data efficiency and performance improvement compared to the PlaNet and AC baselines, and Dreamer V2 \cite{hafner2021mastering} is the first MBRL algorithm to achieve super-human performance with a single GPU in discrete-action Atari benchmarks. The architecture and objective of the LDM of Dreamer are directly inherited from PlaNet's \cite{hafner2019learning}. Dreamer V2 \cite{hafner2021mastering} leverages categorical latent state space representation and \textit{KL balancing} to learn the LDM.

Stochastic Latent Actor-Critic (SLAC) \cite{lee2020stochastic} combines the Soft Actor-Critic's (SAC) \cite{haarnoja2018soft} maximum-entropy RL objective \cite{ziebart2010modeling, levine2018reinforcement} with latent dynamic representation learning to solve \textit{Partially Observable Markov Decision Processes} (POMDP) \cite{kaelbling1998planning}. In contrast to RSSM-based algorithms such as PlaNet and Dreamer, it only learns stochastic latent representation. It exhibits stability, data efficiency and improved performance compared to SAC and PlaNet in several benchmarks \cite{lee2018stochastic} but has never been tested on CDO manipulation tasks. We evaluate the performance of SLAC's policy learning and planning on the LDM in this paper.

Note that we did not examine imitation learning approaches because we mainly focus on reinforcement learning methods in this paper. Nevertheless, Hoque et al. (2022) \cite{hoque2022visuospatial} have shown that VSF outperforms the behaviour-cloning approach DAgger  \cite{seita2020deep} regarding primary metrics.

\section{Method}

We aim to investigate the failure of PlaNet in fabric flattening to develop a model capable of handling this domain. A latent dynamic model (LDM) for pick-and-place (P\&P) fabric-flattening must accurately predict future states based on a sequence of future action trajectories. This allows a planning algorithm to generate a trajectory of candidate actions that minimises a cost function.  We can formulate the model learning problem as a partially observable Markov decision process (POMDP). Our environment is built on SoftGym's \cite{lin2020softgym} cloth-flattening task with P\&P action extension, which comprises 4 parameters $(x_{pick}, y_{pick}, x_{place}, y_{place})$ defined on continuous pixel space [-1, 1] for a single-picker operation \cite{wu2019learning, ma2022learning}. We did not use pick-and-drag action primitive \cite{hoque2022visuospatial, yan2021learning, lin2022learning} for bounding the place position of the fabric on the observation space.

\subsection{Deep Planning Network (PlaNet)} \label{planet}
Recurrent State Space Model (RSSM) \cite{hafner2019learning} is defined under the POMDP setting with the following latent state dynamic: (1) recurrent dynamic model $\bm{h}_t = f(\bm{h}_{t-1}, \bm{z}_{t-1}, \bm{a}_{t-1})$, (2) representation model $\hat{\bm{z}}_t \sim q(\hat{\bm{z}} \; | \; \bm{h}_t, \bm{x}_t)$, and (3) transition predictor $\tilde{\bm{z}}_t \sim p(\tilde{\bm{z}} \;|\; \bm{h}_{t})$, where $\bm{x}$ represents observation, $\bm{a}$ represents action, $\bm{h}$ represents the deterministic latent representation, and $\hat{\bm{z}}$ and $\tilde{\bm{z}}$ represent the prior and posterior stochastic latent states.

PlaNet learns the RSSM to generate accurate observations and rewards from a prior latent distribution for MPC planning. The dynamic model is trained by minimising the KL-divergence between prior and posterior latent states as well as maximising the maximum likelihood of reconstruction of the observation and reward, where it includes an observation predictor $\bm{\hat{x}}_t \sim p(\bm{x} \;|\; \bm{h}_t, \bm{z}_{t})$ and a reward predictor $\hat{r}_t \sim p(r \; |\; \bm{h}_t, \bm{z}_{t})$:

\begin{align}
    &\mathcal{L}_{PlaNet} = \sum_{t=1}^T \Bigg( - 
     \underset{q(\bm{z}_t|\bm{x}_{1:t}, \bm{a}_{1:t})}{\E} \Big[ \log p(\bm{x}_t|\bm{z}_t) + \log p(r_t|\bm{z}_t) \Big] \nonumber\\
    &+   \underset{q(\bm{z}_{t-1}|\bm{x}_{1:t-1}, \bm{a}_{1:t-1})}{\E}\Big[ KL\big[ q(\bm{z}_t|\bm{x}_{1:t}, \bm{a}_{1:{t-1}}) || p(\bm{z}_{t}|\bm{z}_{t-1}, \bm{a}_{t-1}   \big]\Big]
     \Bigg) \label{eqn:planet-objective}
\end{align}

PlaNet adopts mean-square-error (MSE) to learn observation reconstruction and reward prediction from the Gaussian posterior latent space and Kullback–Leibler (KL) divergence for prior learning.

Model predictive control (MPC) is a set of advanced control methods that usually require a learned/known dynamic model to predict the future behaviour of the controlled system  and a cost function to optimise the sampled trajectories. MPC with Cross-Entropy Method (MPC-CEM) is a common variation that samples actions from a multivariate Gaussian distribution and iteratively optimises the distribution's mean and variance from the top trajectories determined by the cost function. PlaNet employs MPC-CEM to produce the policy at run time by unrolling and maximising the accumulative future rewards from the latent prior distributions. It iteratively refines its LDM by exploring the environment and collecting new trajectories generated by the planner.

\subsection{PlaNet-ClothPick} \label{sec:planet-clothpick}

Our PlaNet-ClothPick method is built upon the original PlaNet and trained with our domain-specific reward function (Section \ref{prg:reward}). We train the LDM for cloth-flattening offline using a special data collection script (\ref{prg:data-collection}) so that we bypass the exploration of reinforcement learning, which is a hard problem to address for cloth-like deformable objects \cite{wu2019learning, yan2021learning}. We also adopt \textit{KL balancing} \cite{hafner2021mastering} (Section \ref{prg:kl-balancing}) to enhance latent prior and posterior learning quality. In addition, we apply data augmentation --- observation noise, rotation \cite{lee2021learning}, and vertical flipping --- to improve the learning efficiency and robustness of the method. Finally, we adopt the prior reward learning and the domain-specific planning method ClothMaskPick-MPC (Section \ref{prg:clothmaskpick-mpc}) to further improve manipulation performance. Figure \ref{fig:method} illustrates the further details of the method with its different input/output (I/O) variants.

\subsubsection{Reward Function} \label{prg:reward}
We extend the reward function presented by Hoque et al. (2022) \cite{hoque2022visuospatial}, which is based on  the relative coverage improvement between two consecutive states. We impose penalties as -0.5 for mispicking, large absolute action values (when any equal to or greater than 0.7), and steps that lead to unflattening. Conversely, we assign bonuses as 0.5 to steps that lead to states with high coverage.

\subsubsection{KL balancing} \label{prg:kl-balancing}

In Equation \ref{eqn:planet-objective}, the KL-divergence term aims to learn the prior from the posterior representation and regularises the posterior representation with the prior. To avoid regularising the posterior representation towards poor priors, Dreamer V2 \cite{hafner2021mastering} proposes \textit{KL balancing} that prioritises learning of the prior over regularising the posterior. Combining the two components with an interpolation factor $\alpha=0.8$, \textit{KL balancing} achieves the former by stopping the gradient on the posterior representation and the latter by stopping the gradient on the prior representation. \textit{KL balancing} is a significant factor for improving the asymptotic performance and learning efficiency of Dreamer \cite{hafner2021mastering, hafner2023mastering}.

\subsubsection{ClothMaskPick-MPC} \label{prg:clothmaskpick-mpc}
Building upon original MPC-CEM planning, we restrict the sampling of the first picking action to the cloth mask, which can be extracted from the depth observation of the environment or estimated from the RSSM model if the decoding includes mask prediction. We set the planning horizon to 1, the population of the samples to 5000, and optimisation iterations to 100.

\subsubsection{Data Collection} \label{prg:data-collection}
We generate 1,000 random fabric instances in the SoftGym \cite{lin2020softgym} environment to cover a wide range of shapes and positions. We reserve 100 episodes for assessing the manipulation system; the rest are for developing the method. We also produce 56,100 episodes of 20-step trajectory data from the developing settings for training the LDMs. We delegate 100 episodes for testing the LDM and 56,000 episodes (1.12 million transitional steps) for training the models. 

To cover a wide range of scenarios of P\&P actions on fabrics, we heuristically generate 10\% purely random policy, 10\% corner-biased random policy \cite{hoque2022visuospatial, weng2022fabricflownet}, 40\% Oracle expert flattening and various folding policies, 30\% noisy expert policies and 10\% mix policy for the first 50,000 trajectories. The remaining 6000 trajectories are generated from a highly flattened initial state (above 85\% coverage), where 20\% of the data are produced from  expert flattening policy, 20\% from noisy expert flattening policy and 60\% from cloth-mask small-random-dragging policy.

The manipulation outcome is extremely sensitive to the pick signal relative to the fabric. Oracle expert flattening, expert folding policies and the corner-biased policy are introduced to guide the pick action operating on the corner of the fabric. Noisy expert policies are designed to account for situations where the picking occurs slightly inside or outside the fabric's corners -- within 5\% of errors. The purely random policy addresses picking actions outside of the fabric, while the cloth-mask small-dragging policy is specifically designed for picking on the fabric surface. While most of these policies accommodate a wide range of fabric dragging scenarios, the cloth-mask small-dragging policy is particularly crucial for emphasising small-dragging actions.

The condition of the fabric itself is another crucial factor affecting the operation's outcome. We employ expert folding, noisy expert folding, random folding, and mix policies to include scenarios where the fabric becomes crumpled from a flattened state. The mixed policy is also introduced for diversifying the action types in a single trajectory.

\begin{table}[t]
\centering
\caption{Difficulty tiers regarding normalised coverage of initial states. We allocate 57 of the instances to the corresponding tiers regarding the generated distribution.}
\begin{tabular}{ c|c|c|c } 
 Tier & NC mean ±  std & NC (min, max) & No. Eps\\
  \hline
  0 & 97.64 ± 0 \% & (97.49\%, 97.93\%) & 5\\
 \hline
 1 & 87.78 ± 9.78\% & (73.12\%, 93.50\%) & 4\\
 \hline
 2 & 56.47 ± 3.97\% & (51.82\%, 62.75\%) & 15\\
  \hline
 3 & 40.88 ± 1.58\% & (38.28\%, 43.50\%) & 25\\
  \hline
 4 & 28.39 ± 0.92\% & (27.12\%, 29.53\%) & 8\\
\end{tabular}

\label{tab:tiers}
\end{table}

\begin{figure}[t]%
    \centering
    \subfloat[\centering Reward Study]{{\includegraphics[width=0.4\linewidth]{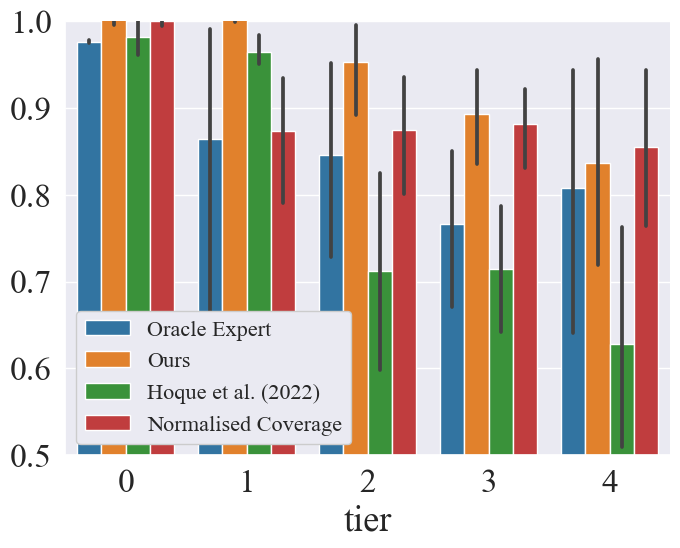} }}%
    \subfloat[\centering Ablation Study]{{\includegraphics[width=0.4\linewidth]{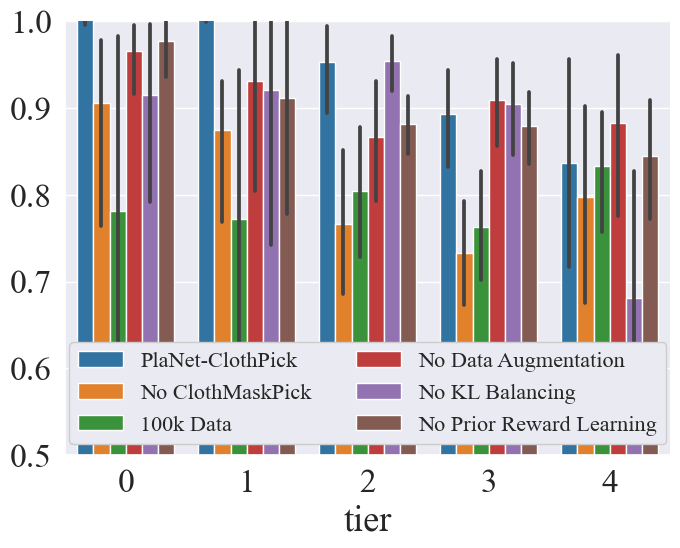} }}%
    \caption{Normalised coverage of PlaNet-ClothPick at step 10 among different tiers. Each constituent element of PlaNet-ClothPick is essential for avoiding unflattening high-coverage articles and reaching higher final coverage. ClothMaskPick-MPC and the specially engineered large dataset are critical for achieving effective flattening in general. Our method even beats the Oracle expert policy used to generate the dataset.}%
    \label{fig:planet-pick-study}%
\end{figure}

\begin{figure}[t]%
    \centering
    \subfloat[\centering Policy Learning Baselines]{{\includegraphics[width=0.4\linewidth]{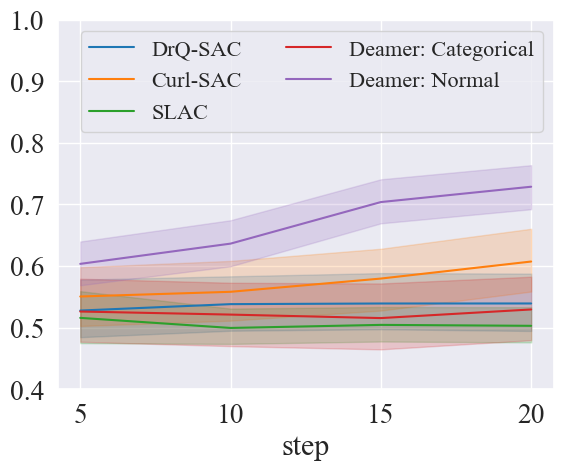} }}%
    \subfloat[\centering Planning Baselines]{{\includegraphics[width=0.4\linewidth]{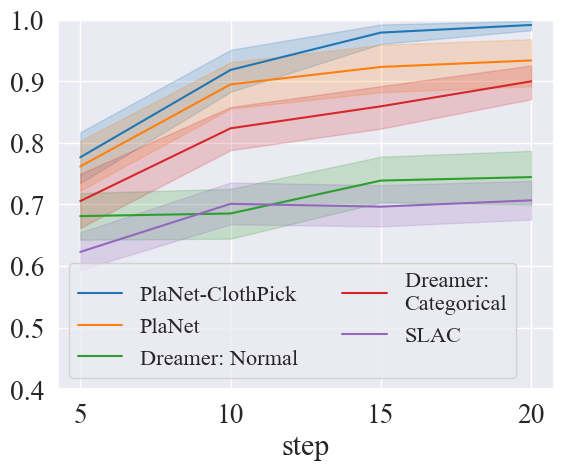} }}%
    \caption{Normalised coverage of different deep reinforcement learning algorithms on pick-and-place fabric flattening. Planning baselines generally shows better performance than policy learning methods on the principal metric, and Planet-ClothPick beats all other state-of-the-art deep reinforcement learning algorithms in fabric-flattening.}%
    \label{fig:DRL-compare}%
\end{figure}

\begin{figure}[t]%
    \centering
    
    \subfloat[\centering Normalised Improvement]{{\includegraphics[width=0.4\linewidth]{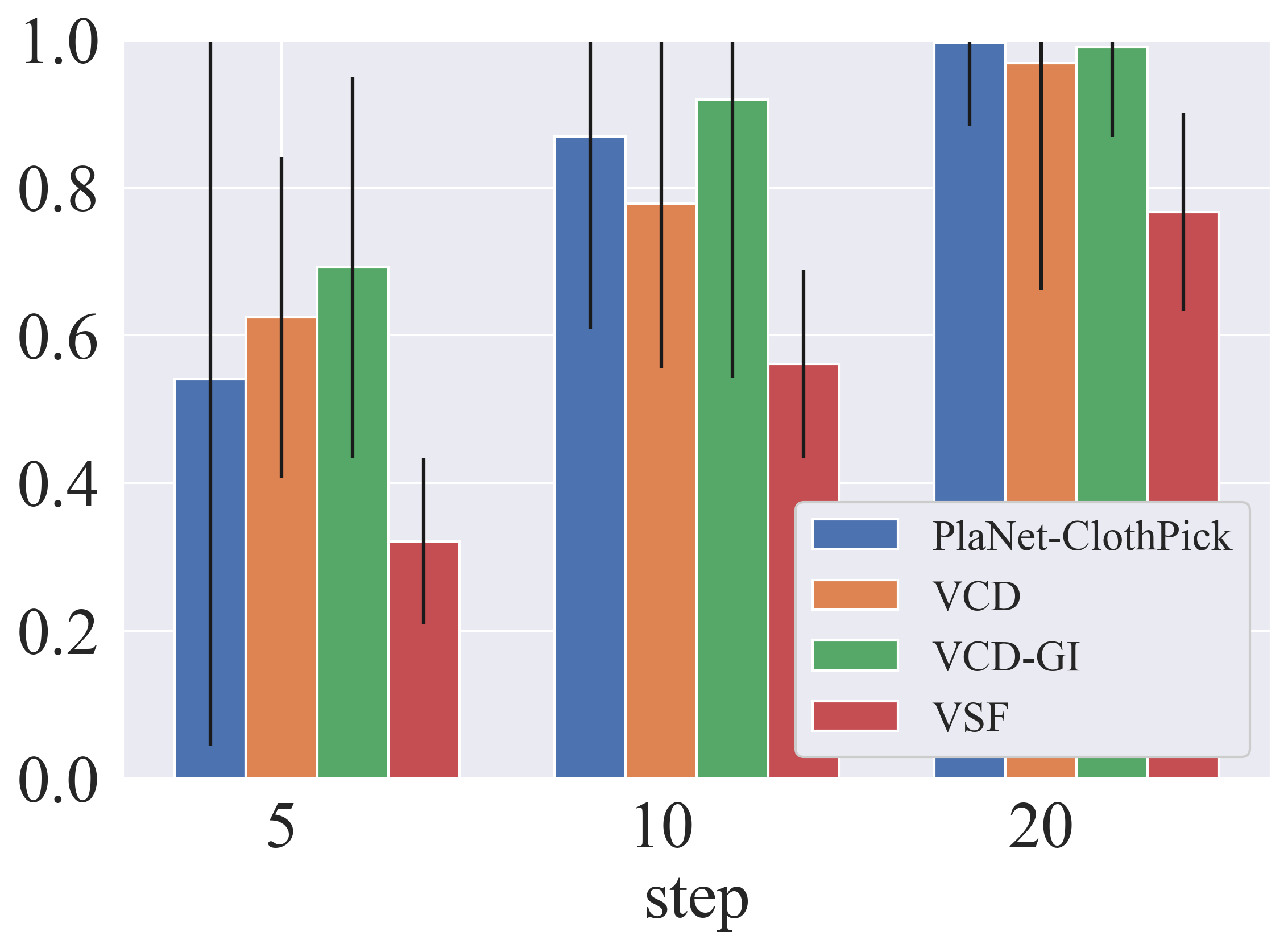} }}%
    \subfloat[\centering  Secondary Metrics ]{{\includegraphics[width=0.4\linewidth]{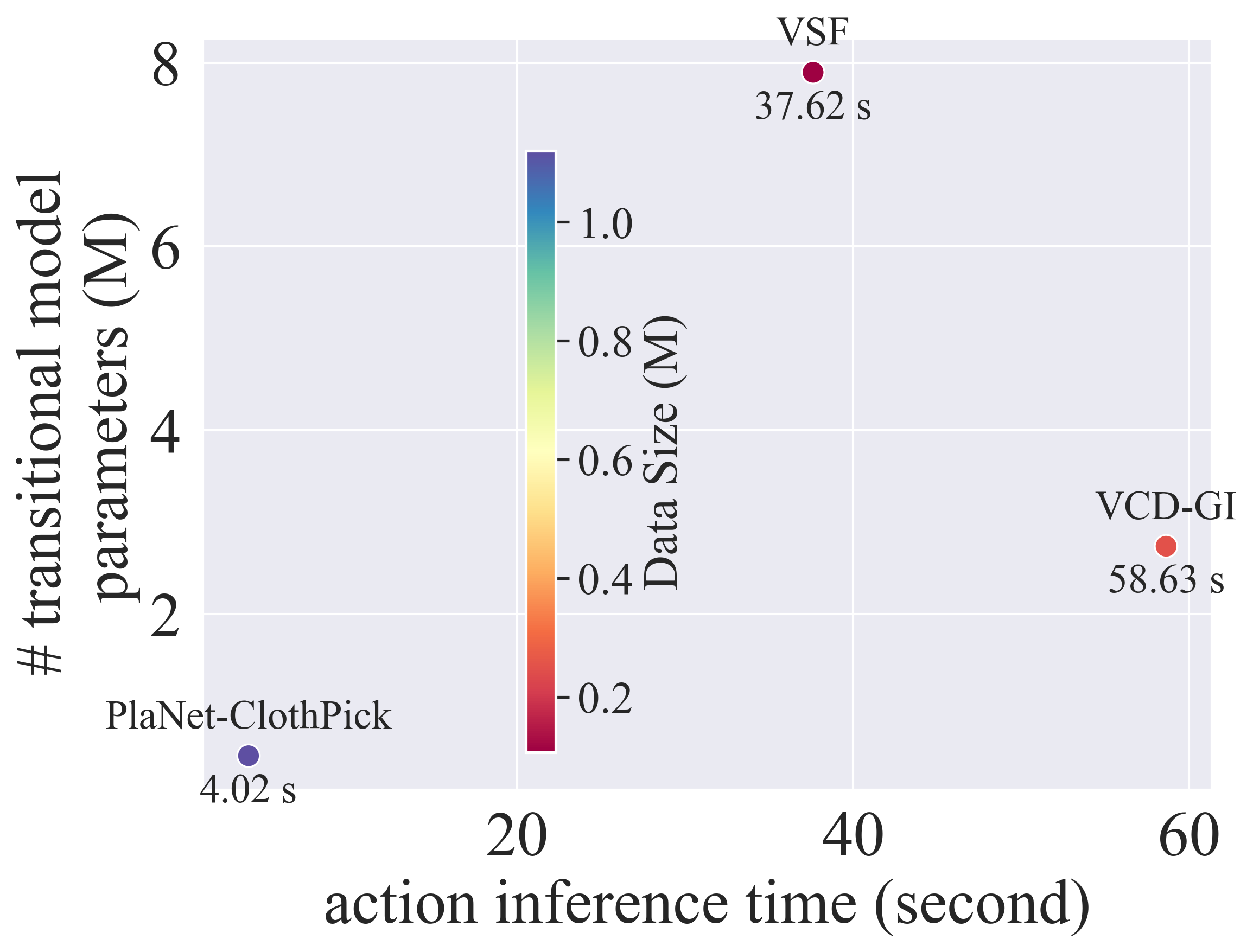} }}%
    \caption{Comparison of PlaNet-ClothPick against state-of-the-art cloth-flattening robotic systems; the colour of the dots in subfigure (b) corresponds to the colour bar for indicating the size of the datasets. Our method achieves a similar level of fabric-fattening as the mesh-based planning methods and surpasses visual planning on the principal metric in simulation, and it showcases a one-order-of-magnitude advantage over the action inference time and transitional model parameters over these systems.}%
    \label{fig:SoTA-flattening-robotic}%
    
\end{figure}

\begin{table*}
\caption{Numerical principal metrics of PlaNet-ClothPick's  input/output variants across different tiers;  we also include the performance of Oracle expert and heuristic method \textit{Wrinkle} \cite{sun2013heuristic, seita2020deep} at the bottom of the table; and the best performance among different steps for all tiers is highlighted with bold text and an asterisk. By default, all these variants use cloth masks fetched from the environment, and we use $\texttt{fm}$ to denote the ones estimated from the model.
The depth-to-depth (D2D) variant performs worst, while other variants perform similarly. Fetching the accurate estimation of the
cloth mask for ClothMaskPick-MPC is important for
selecting the most effective picking position on the fabric.}
\label{tbl:io-variant}
\fontsize{5.8}{9}\selectfont
\centering
\setlength\tabcolsep{2.5pt}
\begin{tabular}{c|c||cccccc||cccccc}
% \toprule

&  & \multicolumn{6}{c||}{Normalised Coverage $\uparrow$} & \multicolumn{6}{c}{Normalised Improvement $\uparrow$}  \\

\hline
 input/output & steps  &  tiers: 0 & 1 & 2 & 3 & 4 & all & 0 &    1 & 2 & 3 & 4 & all \\
 \hline

\cline{1-14}
\multirow{3}{*}{RGB2RGB} & 5  &   1.0 ± 0.01 &   0.93 ± 0.1 &  0.77 ± 0.16 &  0.76 ± 0.14 &  0.63 ± 0.12 &  0.78 ± 0.16 &     \text{1.1 ± 0.36} &  -0.12 ± 1.47 &  0.48 ± 0.34 &  \text{0.59 ± 0.25} &  0.49 ± 0.17 &    \textbf{0.54 ± 0.5}* \\
                                     & 10 &   \text{1.0 ± 0.01} &   \text{1.0 ± 0.01} &  \text{0.95 ± 0.11} &  0.89 ± 0.14 &  0.84 ± 0.18 &  \text{0.92 ± 0.14} &     \text{1.1 ± 0.36} &   1.01 ± 0.03 &  \text{0.89 ± 0.26} &  0.82 ± 0.24 &  0.77 ± 0.26 &   \textbf{0.87 ± 0.26}* \\
                                     & 20 &   \text{1.0 ± 0.01} &   \text{1.0 ± 0.01} &   \text{1.0 ± 0.01} &  0.99 ± 0.02 &  0.96 ± 0.06 &  0.99 ± 0.03 &     1.1 ± 0.36 &   1.01 ± 0.03 &   1.0 ± 0.02 &  0.99 ± 0.04 &  0.95 ± 0.09 &    1.0 ± 0.11 \\

\cline{1-14}
\multirow{3}{*}{D2D} & 5  &  0.99 ± 0.01 &  \text{0.95 ± 0.07} &  0.69 ± 0.15 &  0.69 ± 0.15 &  0.65 ± 0.16 &  0.73 ± 0.17 &     0.48 ± 0.5 &   0.26 ± 0.77 &   0.3 ± 0.34 &  0.47 ± 0.25 &  0.52 ± 0.22 &   0.42 ± 0.35 \\
                                     & 10 &  0.98 ± 0.02 &  0.94 ± 0.07 &  0.84 ± 0.15 &   0.8 ± 0.15 &  0.78 ± 0.14 &  0.83 ± 0.15 &    0.23 ± 0.73 &   0.15 ± 0.87 &  0.62 ± 0.33 &  0.67 ± 0.25 &    0.7 ± 0.2 &   0.59 ± 0.41 \\
                                     & 20 &  0.99 ± 0.02 &  0.96 ± 0.09 &  0.93 ± 0.12 &  0.92 ± 0.12 &  0.86 ± 0.15 &  0.92 ± 0.12 &    0.35 ± 0.78 &   0.51 ± 1.05 &  0.84 ± 0.25 &  0.87 ± 0.19 &   0.8 ± 0.22 &   0.78 ± 0.41 \\

\cline{1-14}
\multirow{3}{*}{RGBD2RGBD} & 5  &  0.98 ± 0.04 &  0.84 ± 0.21 &  0.75 ± 0.18 &  0.74 ± 0.19 &  0.68 ± 0.08 &  0.76 ± 0.18 &    0.37 ± 1.58 &  -0.02 ± 0.53 &   0.42 ± 0.4 &  0.57 ± 0.31 &  0.56 ± 0.11 &   0.47 ± 0.55 \\
                                     & 10 &   \text{1.0 ± 0.01} &  0.98 ± 0.02 &  0.87 ± 0.19 &  0.92 ± 0.14 &  0.87 ± 0.15 &  0.91 ± 0.15 &    1.01 ± 0.52 &    \text{0.74 ± 0.3} &  0.68 ± 0.47 &  0.86 ± 0.24 &  0.82 ± 0.21 &   0.81 ± 0.35 \\
                                     & 20 &   \text{1.0 ± 0.01} &  0.99 ± 0.01 &  0.99 ± 0.03 &  0.99 ± 0.04 &  0.98 ± 0.04 &  0.99 ± 0.03 &    1.01 ± 0.52 &   0.87 ± 0.18 &  0.99 ± 0.06 &  0.99 ± 0.07 &  0.98 ± 0.05 &   0.98 ± 0.16 \\

\cline{1-14}
\multirow{3}{*}{RGB2Mask} & 5  &  0.99 ± 0.01 &  \text{0.95 ± 0.03} &  0.78 ± 0.19 &  0.76 ± 0.17 &  0.68 ± 0.19 &  \textbf{0.79 ± 0.18}* &    0.46 ± 0.47 &   \text{0.32 ± 0.53} &  0.51 ± 0.41 &   0.6 ± 0.29 &  0.56 ± 0.26 &   \text{0.54 ± 0.35}* \\
                                     & 10 &  0.99 ± 0.01 &  0.96 ± 0.06 &   0.94 ± 0.1 &  \text{0.92 ± 0.11} &  0.93 ± 0.08 &   \textbf{0.94 ± 0.1}* &    0.39 ± 0.62 &   0.45 ± 0.92 &  0.88 ± 0.22 &  \text{0.87 ± 0.19} &  0.91 ± 0.11 &   0.81 ± 0.36 \\
                                     & 20 &   \text{1.0 ± 0.01} &  0.98 ± 0.04 &   1.0 ± 0.04 &  \text{1.0 ± 0.02} &  0.99 ± 0.03 &   \textbf{1.0 ± 0.03}* &    1.04 ± 0.37 &   0.66 ± 0.57 &   1.0 ± 0.09 &   \text{1.0 ± 0.04} &  0.98 ± 0.04 &   0.97 ± 0.19 \\
                                     
\cline{1-14}
\multirow{3}{*}{RGB2Mask-fm} & 5  &  0.99 ± 0.02 &  0.94 ± 0.07 &   \text{0.8 ± 0.18} &   0.7 ± 0.16 &   0.67 ± 0.1 &  0.77 ± 0.17 &    0.34 ± 0.84 &   0.15 ± 1.07 &  \text{0.55 ± 0.41} &   0.5 ± 0.27 &  0.53 ± 0.14 &   0.48 ± 0.45 \\
                                     & 10 &   \text{1.0 ± 0.01} &  0.96 ± 0.04 &   0.94 ± 0.1 &  0.88 ± 0.16 &   \text{0.94 ± 0.1} &  0.92 ± 0.13 &     0.9 ± 0.42 & 0.68 ± 0.32 &  0.85 ± 0.23 &   0.8 ± 0.27 &  \text{0.92 ± 0.13} &   0.83 ± 0.26 \\
                                     & 20 &   \text{1.0 ± 0.01} &  0.98 ± 0.03 &  0.98 ± 0.06 &  0.98 ± 0.05 &   1.0 ± 0.02 &  0.98 ± 0.05 &     0.9 ± 0.42 &   0.77 ± 0.33 &  0.95 ± 0.13 &  0.96 ± 0.09 &   1.0 ± 0.03 &   0.94 ± 0.17 \\

\cline{1-14}
\multirow{3}{*}{D2Mask} & 5  &   0.8 ± 0.27 &  0.91 ± 0.09 &   0.76 ± 0.2 &  0.72 ± 0.18 &  \text{0.69 ± 0.25} &   0.75 ± 0.2 &  -7.12 ± 11.33 &  -0.29 ± 1.17 &  0.44 ± 0.49 &  0.52 ± 0.31 &  \text{0.57 ± 0.35} &  -0.22 ± 3.75 \\
                                     & 10 &  0.89 ± 0.25 &  0.88 ± 0.25 &  0.83 ± 0.18 &  0.83 ± 0.21 &   0.85 ± 0.2 &   0.84 ± 0.2 &   -3.37 ± 9.89 &   -0.53 ± 3.1 &  0.61 ± 0.43 &  0.71 ± 0.35 &   0.8 ± 0.29 &    0.25 ± 3.0 \\
                                     & 20 &   \text{1.0 ± 0.01} &  0.99 ± 0.04 &   1.0 ± 0.02 &  0.98 ± 0.08 &  \text{1.01 ± 0.01} &  0.99 ± 0.06 &    0.99 ± 0.62 &    0.8 ± 0.44 &   1.0 ± 0.05 &  0.96 ± 0.14 &  \text{1.01 ± 0.01} &   0.97 ± 0.22 \\
\cline{1-14}                                     
\multirow{3}{*}{D2Mask-fm} & 5  &  0.87 ± 0.15 &   0.94 ± 0.1 &  0.65 ± 0.17 &  0.63 ± 0.18 &  \text{0.69 ± 0.15} &  0.69 ± 0.19 &   -5.03 ± 7.15 &   0.13 ± 1.57 &  0.18 ± 0.42 &  0.37 ± 0.31 &  \text{0.57 ± 0.22} &   -0.14 ± 2.5 \\
                                     & 10 &  0.74 ± 0.26 &  0.97 ± 0.08 &  0.84 ± 0.14 &  0.81 ± 0.18 &  0.81 ± 0.18 &  0.82 ± 0.17 &  -10.5 ± 10.75 &   0.46 ± 1.13 &  0.63 ± 0.35 &   0.68 ± 0.3 &  0.74 ± 0.25 &  -0.32 ± 4.31 \\
                                     & 20 &  0.97 ± 0.06 &  \text{1.01 ± 0.01} &  0.94 ± 0.12 &   0.9 ± 0.16 &  0.92 ± 0.11 &  0.93 ± 0.13 &   -0.25 ± 3.07 &   \text{1.08 ± 0.07} &  0.85 ± 0.28 &  0.83 ± 0.28 &  0.89 ± 0.15 &   0.77 ± 0.91 \\

\cline{1-14}
\multirow{3}{*}{D2RGB} & 5  &  0.96 ± 0.06 &  \text{0.95 ± 0.06} &  0.77 ± 0.19 &  0.69 ± 0.17 &  0.59 ± 0.16 &  0.74 ± 0.19 &   -0.55 ± 2.75 &   0.27 ± 0.83 &  0.47 ± 0.43 &  0.47 ± 0.28 &  0.43 ± 0.23 &   0.36 ± 0.87 \\
                                     & 10 &  0.99 ± 0.03 &   \text{1.0 ± 0.01} &  0.91 ± 0.14 &  0.89 ± 0.14 &  0.83 ± 0.12 &   0.9 ± 0.13 &    0.73 ± 1.13 &   0.94 ± 0.08 &  0.79 ± 0.32 &  0.82 ± 0.24 &  0.76 ± 0.17 &    0.8 ± 0.38 \\
                                     & 20 &   \text{1.0 ± 0.01} &   1.0 ± 0.01 &  \text{1.01 ± 0.01} &  0.98 ± 0.08 &   1.0 ± 0.01 &   \textbf{1.0 ± 0.05}* &     \text{1.2 ± 0.45} &   0.95 ± 0.08 &  \text{1.01 ± 0.02} &  0.97 ± 0.13 &  \text{1.01 ± 0.02} &   \textbf{1.01 ± 0.16}* \\
\hline
\hline
\cline{1-14}
\multirow{3}{*}{Oracle Expert} & 5  &   0.98 ± 0.0 &  0.82 ± 0.23 &  0.73 ± 0.25 &   0.62 ± 0.2 &  0.52 ± 0.14 &  0.68 ± 0.23 &  -0.03 ± 0.18 &   -1.3 ± 3.12 &   0.4 ± 0.55 &  0.35 ± 0.35 &   0.34 ± 0.2 &   0.21 ± 0.92 \\
        & 10 &   0.98 ± 0.0 &  0.86 ± 0.25 &  0.85 ± 0.24 &  0.77 ± 0.24 &  0.81 ± 0.25 &  0.82 ± 0.23 &  -0.01 ± 0.17 &  -0.77 ± 3.27 &  0.66 ± 0.53 &   0.61 ± 0.4 &  0.73 ± 0.35 &   0.49 ± 0.94 \\
        & 20 &   0.98 ± 0.0 &  0.98 ± 0.01 &  0.92 ± 0.19 &  0.89 ± 0.22 &  0.97 ± 0.05 &  0.92 ± 0.18 &   0.03 ± 0.17 &   0.81 ± 0.13 &   0.82 ± 0.4 &  0.81 ± 0.38 &  0.96 ± 0.07 &    0.77 ± 0.4 \\
\cline{1-14}
\multirow{3}{*}{Wrinkle} & 5  &  0.98 ± 0.0 &  0.87 ± 0.12 &  0.69 ± 0.16 &  0.58 ± 0.13 &  0.52 ± 0.14 &  0.65 ± 0.19 &     0.0 ± 0.0 &  -0.83 ± 2.08 &  0.29 ± 0.36 &  0.28 ± 0.22 &  0.33 ± 0.19 &  0.19 ± 0.61 \\
        & 10 &  0.98 ± 0.0 &   0.9 ± 0.09 &  0.75 ± 0.14 &  0.63 ± 0.17 &  0.58 ± 0.14 &   0.7 ± 0.19 &     0.0 ± 0.0 &  -0.37 ± 1.34 &  0.43 ± 0.32 &  0.37 ± 0.29 &   0.42 ± 0.2 &  0.31 ± 0.46 \\
        & 20 &  0.98 ± 0.0 &  0.92 ± 0.07 &  0.88 ± 0.12 &   0.7 ± 0.17 &   0.68 ± 0.2 &  0.78 ± 0.18 &     0.0 ± 0.0 &  -0.14 ± 1.02 &  0.73 ± 0.28 &   0.48 ± 0.3 &  0.55 ± 0.27 &  0.47 ± 0.43 \\

% \cline{1-5}
% \cline{2-5}
\bottomrule
\end{tabular}
\end{table*}

\begin{figure*}[t]

\subfloat[\centering Depth and Mask Reconstruction of PlaNet-ClothPick]{{\includegraphics[width=0.49\linewidth]{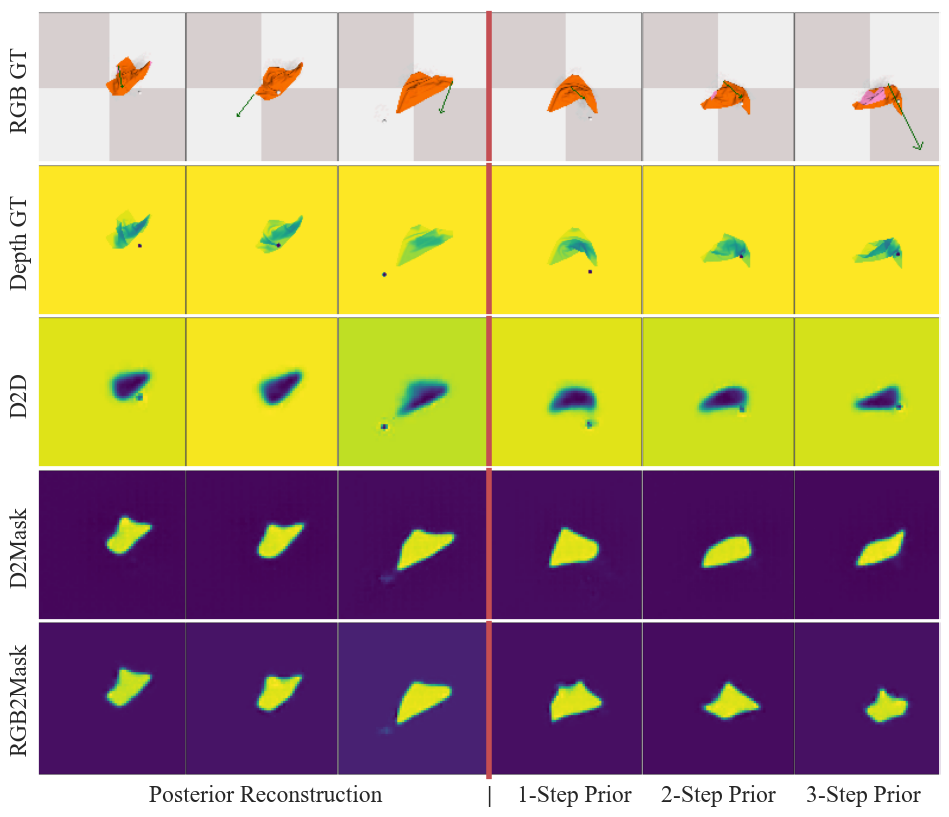}}}
\subfloat[\centering RGB2RGB Reconstruction of LDMs]{{\includegraphics[width=0.5\linewidth]{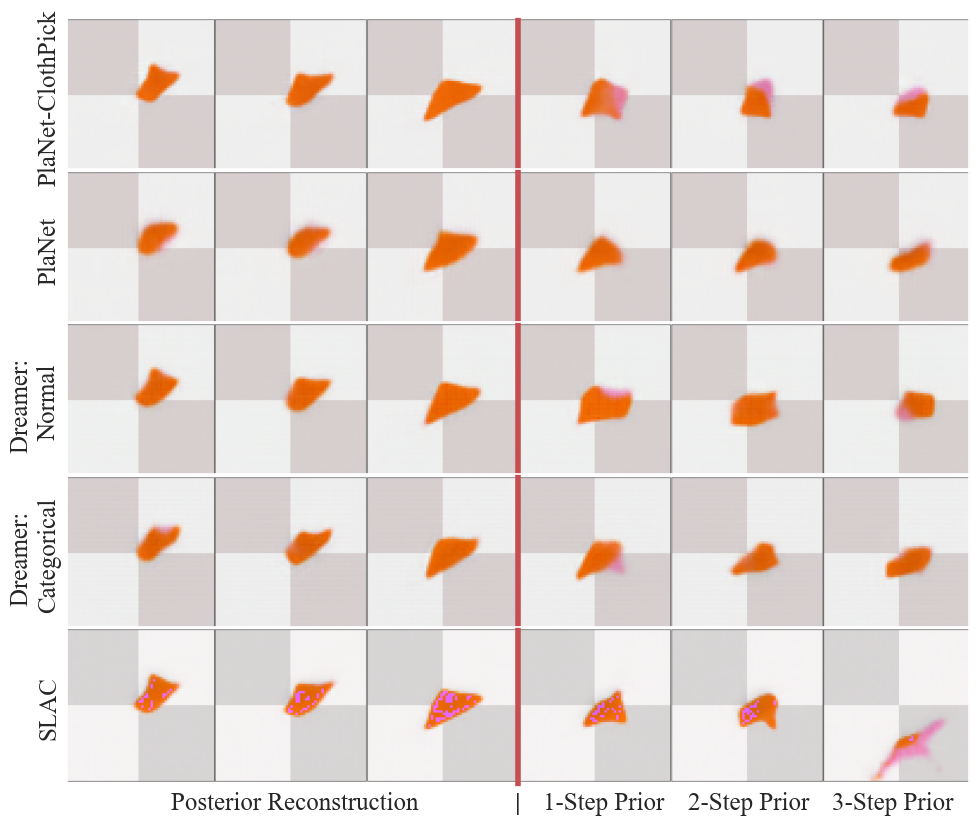}}}
    \caption{Reconstruction quality of latent dynamic models compared to ground truth (GT). By incorporating \textit{KL balancing}, PlaNet-ClothPick produces the best posterior and prior observation reconstruction quality.}
    \label{fig:reconstruction}
\end{figure*}

\section{Experiments}

We standardise the pick-and-place (P\&P) fabric-flattening simulated environment and conduct all experiments in SoftGym \cite{lin2020softgym}, which originally provided the basic functionality of the simulation and the performance benchmark on the velocity control. We assess manipulation performance through normalised coverage (NC) and normalised improvement (NI) across the action steps up to 20 steps, as it is the standard in the cloth-flattening literature. We additionally evaluate latent dynamic models (LDMs) through the observation reconstruction from posteriors, which gives a good indication of posterior representation learning, which is essential for providing good initial latent states for the planning (see Figure \ref{fig:reconstruction}). Following VSF \cite{hoque2022visuospatial}, we manually select a subset of testing states and classify them into five tiers based on initial coverage (see Table \ref{tab:tiers}). Note that all the methods in this work are trained offline as other SoTA robotic systems \cite{hoque2022visuospatial, ma2022learning, lin2022learning}.

\subsection{Comparison against SoTA methods} \label{general-study}
We benchmark PlaNet-ClothPick (Section \ref{sec:planet-clothpick}) on fabric flattening with our reward function against SoTA policy learning deep reinforcement learning (DRL) algorithms, such as Curl-SAC \cite{laskin2020curl}, DrQ-SAC \cite{kostrikov2020image}, Dreamer with normal and categorical distributions, and SLAC (Figure \ref{fig:DRL-compare}(a)), as well as ClothMaskPick-MPC (Section \ref{prg:clothmaskpick-mpc}) planning on the LDMs from PlaNet, Dreamer with both variants and SLAC (Figure \ref{fig:DRL-compare}(b)). Note that the policy learning DRL baselines are trained with 500,000 update steps, and LDMs with 100,000 update steps --- we employ ClothMaskPick-MPC to plan on these LDMs for consistency. Both sets of baselines learn from our special dataset with the proposed reward function and $64\times 64$ RGB images as input/output (I/O) observation by rescaling the values within the range of [-0.5, 0.5]. For controlling the variables while comparing against PlaNet-ClothPick, only the LDM baselines apply our data augmentation, as some of the policy learning methods come with their own.

We also compare our method against the reported performance of the previous SoTA cloth-flattening robotic systems (Figure \ref{fig:SoTA-flattening-robotic}), such as VSF, VCD and VCD-GI,  from Lin et al. (2022) \cite{lin2022learning}. In addition, we experiment on the heuristic method \textit{Wrinkle} proposed by Sun et al. (2013) \cite{sun2013heuristic} via integrating the corresponding implementation of Seita et al. (2020) \cite{seita2020deep} to our environment. Note that this implementation approximates the detection of wrinkles from the true particle positions of the fabric rather than from the depth camera as in the original method; hence, it only works in simulation.  Then, we examine the action inference time of all successful methods on a GeForce RTX 3090 GPU with the systems' default setting.

Our result shows that PlaNet-ClothPick outperforms all the general DRL algorithms in fabric flattening. It also statistically surpasses SoTA NI-against-step performance of VSF and VCD, reaching the same level of competence as VCD-GI. Moreover, our method exhibits around $10\times$ faster action inference time and $10\times$ fewer transitional model parameters compared to the three baselines. However, it does need $10\times$ more data to train.

\subsection{Study on PlaNet-ClothPick} \label{component-study} \label{pick-readjust}

\emph{(i) How significant are the different components of the PlaNet-ClothPick?}

We train our model on two other reward functions: normalised coverage and the reward from Hoque et al. (2022) \cite{hoque2022visuospatial}. Figure \ref{fig:planet-pick-study}(a) indicates that our reward function is key to achieving near-perfect performance, especially for cases with low initial coverage. It also shows that PlaNet-ClothPick outperforms the expert policy used for data generation.

Figure \ref{fig:planet-pick-study}(b) presents the significance of the different parts of the model during the training and inference time. We observe that cross-entropy model predictive control (MPC-CEM) often produces an action slightly outside the cloth, which always misses the cloth and makes the algorithm operationally inefficient (see Figure \ref{fig:traj-compare}); the proposed ClothMaskPick-MPC is critical for achieving effective flattening in general. Sufficient data, data augmentation, \textit{KL balancing} and prior reward learning are all essential for PlaNet-ClothPick to achieve near-perfect flattening.

\begin{figure}[t]
    \centering
    \includegraphics[width=0.6\linewidth]{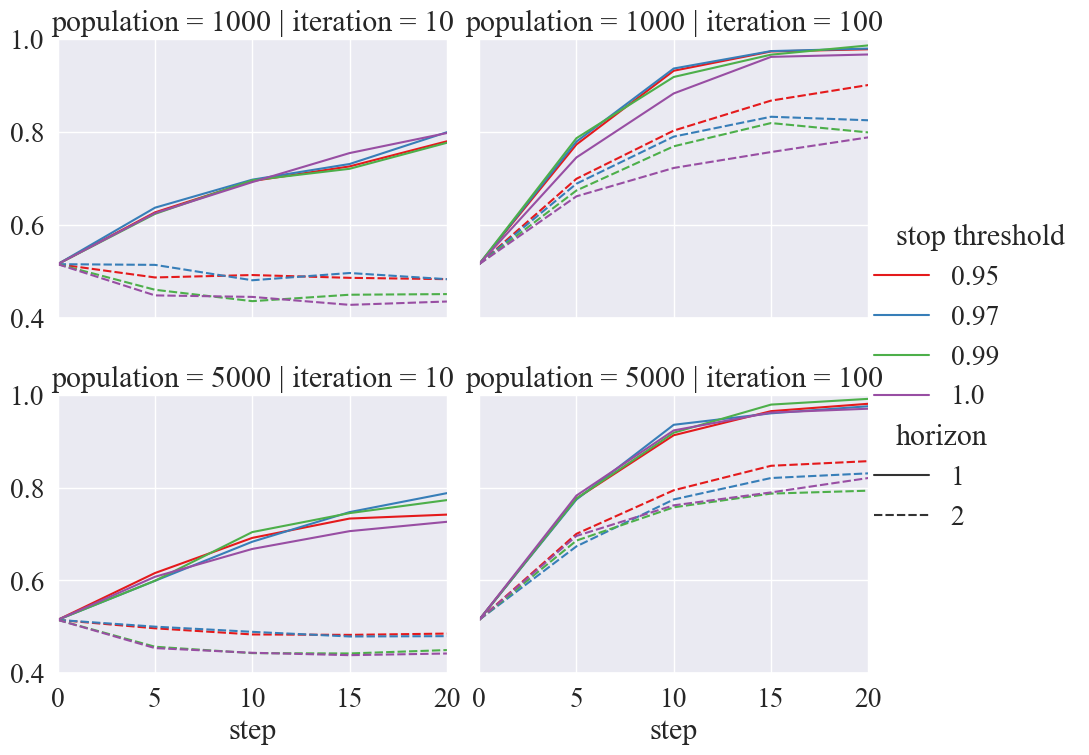}
    \caption{Normalised coverage of ClothMaskPick-MPC with different hyperparameters. More optimisation iterations and populations produce better and more effective planning results. However, the proposed planning method struggles with multi-step horizons, as it cannot estimate the prior cloth mask for further planning steps. }
    \label{fig:clothmask-mpc-study}

\end{figure}

\begin{figure}[t]%
    \centering
    \subfloat[\centering Prediction Error]{{\includegraphics[width=0.4\linewidth]{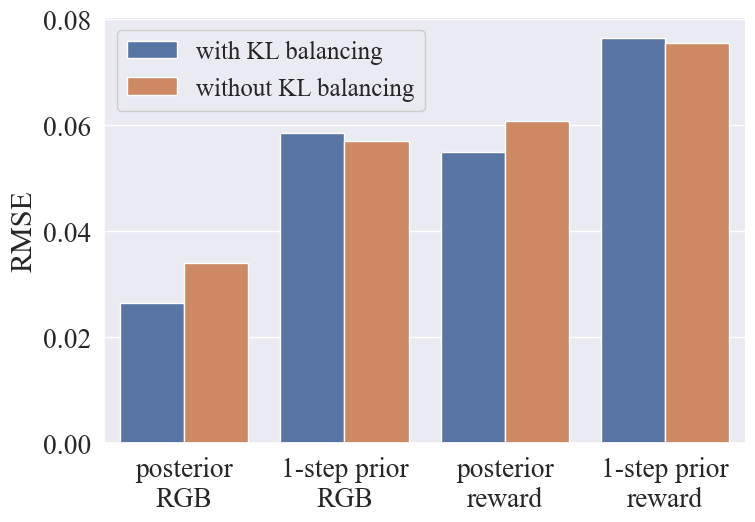} }}%
    \subfloat[\centering Latent Properties]{{\includegraphics[width=0.39\linewidth]{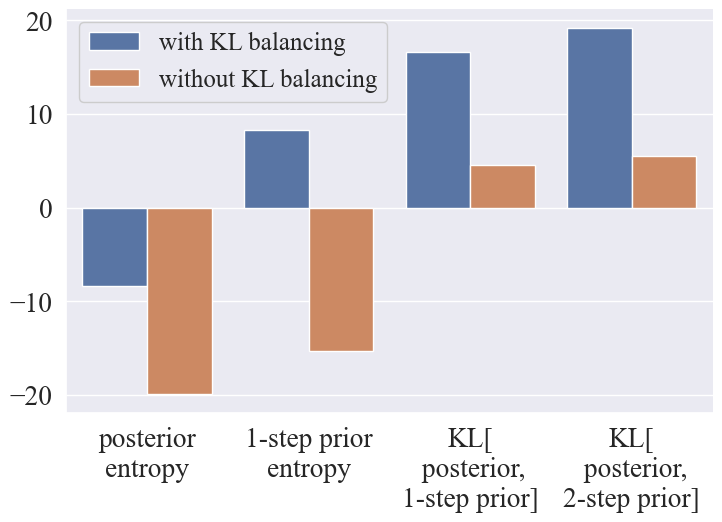} }}%
        \caption{Effects of \textit{KL balancing}. Although \textit{KL balancing} increases the entropy of the latent space and the divergence between the posterior and prior representation, it produces better latent space reflected by the better accuracy of the observation and reward posterior prediction. }
        
    \label{fig:kl-balancing}%
\end{figure}

\emph{(ii)}  \textit{How does KL balancing contribute to the success of the PlaNet-ClothPick?}
Figure \ref{fig:kl-balancing} shows that \textit{KL balancing} generates a latent space that leads to better observation and reward prediction accuracy, which provides a better initial state estimate for planning. 

\emph{(iii) How does the RSSM model's input/output variants affect the performance of PlaNet-ClothPick?}

Table \ref{tbl:io-variant} indicates that the D2D variant performs worse than the D2RGB and D2Mask variants; combining the reconstruction quality of the variants from Figure \ref{fig:reconstruction} suggests that compressing the depth-only information cannot produce good latent representation for achieving fabric-flattening. Besides, the depth or RGB input of the RSSM model does not affect the performance of PlaNet-ClothPick when the output observation is informative enough for learning good latent representation.

In addition, comparing the results of RGB2Mask to RGB2Mask-fm and D2Mask to D2Mask-fm, we conclude that fetching the accurate estimation of the cloth mask for ClothMaskPick-MPC is important for selecting the most effective picking position on the cloth.
 
\emph{(iii) How robust is ClothMaskPick-MPC?}

We examine the hyper-parameters of ClothMaskPick-MPC on the RGB2RGB variant of PlaNet-ClothPick. Figure \ref{fig:clothmask-mpc-study} demonstrates that more optimisation iterations and larger populations produce better and more effective planning results. However, the proposed planning method struggles with multi-step horizons, as it cannot estimate the prior cloth mask for further planning steps.

\section{Conclusion}
This paper investigates the failure of latent dynamic models (LDMs) on fabric flattening. To our knowledge, this is the first time an Recurrent State Space Models (RSSM) based model has shown state-of-the-art (SoTA) performance on the fabric-flattening task. We find that the sharp discontinuity of the transition function on the fabric's contour makes it difficult to learn an accurate LDM, causing the Model Predictive Control (MPC) planner to produce pick actions slightly outside of the cloth.  We employ ClothMaskPick-MPC, \textit{KL balancing}, prior reward learning, data augmentation, and special data collection to improve the performance and robustness of PlaNet in this domain.

Our mesh-free method PlaNet-ClothPick achieves SoTA performance regarding primary metrics among all the reinforcement learning methods, an order-of-magnitude advantage over the action inference time and the number of transitional model parameters compared to the previous SoTA robotic systems in this domain.

In the future, we would like to investigate our method in real-world trials and extend its application to garment flattening. We also plan to reduce the inductive biases we introduced in the data collection by applying SoTA exploration strategies and those in planning algorithms by combining policy learning and planning. Finally, we will investigate the multi-step prediction ability of the RSSM models and make the planning algorithm more robust with multi-step planning, potentially making the flattening more operationally effective.

% % \addtolength{\textheight}{-6cm}   % This command serves to balance the column lengths
% %                                   % on the last page of the document manually. It shortens
% %                                   % the textheight of the last page by a suitable amount.
% %                                   % This command does not take effect until the next page
% %                                   % so it should come on the page before the last. Make
% %                                   % sure that you do not shorten the textheight too much.

% %%%%%%%%%%%%%%%%%%%%%%%%%%%%%%%%%%%%%%%%%%%%%%%%%%%%%%%%%%%%%%%%%%%%%%%%%%%%%%%%

% %%%%%%%%%%%%%%%%%%%%%%%%%%%%%%%%%%%%%%%%%%%%%%%%%%%%%%%%%%%%%%%%%%%%%%%%%%%%%%%%

% %%%%%%%%%%%%%%%%%%%%%%%%%%%%%%%%%%%%%%%%%%%%%%%%%%%%%%%%%%%%%%%%%%%%%%%%%%%%%%%%

%%%%%%%%%%%%%%%%%%%%%%%%%%%%%%%%%%%%%%%%%%%%%%%%%%%%%%%%%%%%%%%%%%%%%%%%%%%%%%%%
\newpage
\bibliographystyle{IEEEtran}
\bibliography{ref}

\begin{thebibliography}{10}
\providecommand{\url}[1]{#1}
\csname url@samestyle\endcsname
\providecommand{\newblock}{\relax}
\providecommand{\bibinfo}[2]{#2}
\providecommand{\BIBentrySTDinterwordspacing}{\spaceskip=0pt\relax}
\providecommand{\BIBentryALTinterwordstretchfactor}{4}
\providecommand{\BIBentryALTinterwordspacing}{\spaceskip=\fontdimen2\font plus
\BIBentryALTinterwordstretchfactor\fontdimen3\font minus \fontdimen4\font\relax}
\providecommand{\BIBforeignlanguage}[2]{{%
\expandafter\ifx\csname l@#1\endcsname\relax
\typeout{** WARNING: IEEEtran.bst: No hyphenation pattern has been}%
\typeout{** loaded for the language `#1'. Using the pattern for}%
\typeout{** the default language instead.}%
\else
\language=\csname l@#1\endcsname
\fi
#2}}
\providecommand{\BIBdecl}{\relax}
\BIBdecl

\bibitem{hafner2019learning}
D.~Hafner, T.~Lillicrap, I.~Fischer, R.~Villegas, D.~Ha, H.~Lee, and J.~Davidson, ``Learning latent dynamics for planning from pixels,'' in \emph{International Conference on Machine Learning}.\hskip 1em plus 0.5em minus 0.4em\relax Long Beach, CA, USA: PMLR, 10--15 June 2019, pp. 2555--2565.

\bibitem{hafner2020dream}
\BIBentryALTinterwordspacing
D.~Hafner, T.~Lillicrap, J.~Ba, and M.~Norouzi, ``Dream to control: Learning behaviors by latent imagination,'' in \emph{International Conference on Learning Representations}, Addis Ababa, Ethiopia, 26--30 April 2020. [Online]. Available: \url{https://openreview.net/forum?id=S1lOTC4tDS}
\BIBentrySTDinterwordspacing

\bibitem{hafner2021mastering}
\BIBentryALTinterwordspacing
D.~Hafner, T.~P. Lillicrap, M.~Norouzi, and J.~Ba, ``Mastering atari with discrete world models,'' in \emph{International Conference on Learning Representations}, 2021. [Online]. Available: \url{https://openreview.net/forum?id=0oabwyZbOu}
\BIBentrySTDinterwordspacing

\bibitem{hafner2023mastering}
D.~Hafner, J.~Pasukonis, J.~Ba, and T.~Lillicrap, ``Mastering diverse domains through world models,'' \emph{arXiv preprint arXiv:2301.04104}, 2023.

\bibitem{tassa2018deepmind}
Y.~Tassa, Y.~Doron, A.~Muldal, T.~Erez, Y.~Li, D.~d.~L. Casas, D.~Budden, A.~Abdolmaleki, J.~Merel, A.~Lefrancq \emph{et~al.}, ``Deepmind control suite,'' \emph{arXiv preprint arXiv:1801.00690}, 2018.

\bibitem{bellemare2013arcade}
M.~G. Bellemare, Y.~Naddaf, J.~Veness, and M.~Bowling, ``The arcade learning environment: An evaluation platform for general agents,'' \emph{Journal of Artificial Intelligence Research}, vol.~47, pp. 253--279, 2013.

\bibitem{sun2013heuristic}
L.~Sun, G.~Aragon-Camarasa, P.~Cockshott, S.~Rogers, and J.~P. Siebert, ``A heuristic-based approach for flattening wrinkled clothes,'' in \emph{Conference Towards Autonomous Robotic Systems}.\hskip 1em plus 0.5em minus 0.4em\relax Oxford, UK: Springer, 8–30 August 2013, pp. 148--160.

\bibitem{seita2020deep}
D.~Seita, A.~Ganapathi, R.~Hoque, M.~Hwang, E.~Cen, A.~K. Tanwani, A.~Balakrishna, B.~Thananjeyan, J.~Ichnowski, N.~Jamali \emph{et~al.}, ``Deep imitation learning of sequential fabric smoothing from an algorithmic supervisor,'' in \emph{2020 IEEE/RSJ International Conference on Intelligent Robots and Systems (IROS)}.\hskip 1em plus 0.5em minus 0.4em\relax IEEE, 2020, pp. 9651--9658.

\bibitem{hoque2022visuospatial}
R.~Hoque, D.~Seita, A.~Balakrishna, A.~Ganapathi, A.~K. Tanwani, N.~Jamali, K.~Yamane, S.~Iba, and K.~Goldberg, ``Visuospatial foresight for physical sequential fabric manipulation,'' \emph{Autonomous Robots}, vol.~46, no.~1, pp. 175--199, 2022a.

\bibitem{kadi2023data}
H.~A. Kadi and K.~Terzi{\'c}, ``Data-driven robotic manipulation of cloth-like deformable objects: The present, challenges and future prospects,'' \emph{Sensors}, vol.~23, no.~5, p. 2389, 2023.

\bibitem{seita2021learning}
D.~Seita, P.~Florence, J.~Tompson, E.~Coumans, V.~Sindhwani, K.~Goldberg, and A.~Zeng, ``Learning to rearrange deformable cables, fabrics, and bags with goal-conditioned transporter networks,'' in \emph{2021 IEEE International Conference on Robotics and Automation (ICRA)}.\hskip 1em plus 0.5em minus 0.4em\relax Xi’an, China: IEEE, 30 May--5 June 2021, pp. 4568--4575.

\bibitem{wu2019learning}
Y.~Wu, W.~Yan, T.~Kurutach, L.~Pinto, and P.~Abbeel, ``Learning to manipulate deformable objects without demonstrations,'' \emph{arXiv preprint arXiv:1910.13439}, 2019.

\bibitem{ma2022learning}
X.~Ma, D.~Hsu, and W.~S. Lee, ``Learning latent graph dynamics for visual manipulation of deformable objects,'' in \emph{2022 International Conference on Robotics and Automation (ICRA)}.\hskip 1em plus 0.5em minus 0.4em\relax IEEE, 2022, pp. 8266--8273.

\bibitem{lin2022learning}
X.~Lin, Y.~Wang, Z.~Huang, and D.~Held, ``Learning visible connectivity dynamics for cloth smoothing,'' in \emph{Conference on Robot Learning}.\hskip 1em plus 0.5em minus 0.4em\relax Auckland, New Zealand: PMLR, 5--18 December 2022, pp. 256--266.

\bibitem{huang2022mesh}
Z.~Huang, X.~Lin, and D.~Held, ``Mesh-based dynamics with occlusion reasoning for cloth manipulation,'' \emph{arXiv preprint arXiv:2206.02881}, 2022.

\bibitem{xu2022dextairity}
Z.~Xu, C.~Chi, B.~Burchfiel, E.~Cousineau, S.~Feng, and S.~Song, ``Dextairity: Deformable manipulation can be a breeze,'' in \emph{Proceedings of Robotics: Science and Systems (RSS)}, New York, NY, USA, 27 June--1 July 2022.

\bibitem{ha2022flingbot}
H.~Ha and S.~Song, ``Flingbot: The unreasonable effectiveness of dynamic manipulation for cloth unfolding,'' in \emph{Conference on Robot Learning}.\hskip 1em plus 0.5em minus 0.4em\relax Auckland, New Zealand: PMLR, 15--18 Decemeber 2022, pp. 24--33.

\bibitem{yan2021learning}
W.~Yan, A.~Vangipuram, P.~Abbeel, and L.~Pinto, ``Learning predictive representations for deformable objects using contrastive estimation,'' in \emph{Conference on Robot Learning}.\hskip 1em plus 0.5em minus 0.4em\relax London, UK: PMLR, 8--11 November 2021, pp. 564--574.

\bibitem{lin2020softgym}
X.~Lin, Y.~Wang, J.~Olkin, and D.~Held, ``Softgym: Benchmarking deep reinforcement learning for deformable object manipulation,'' in \emph{Conference on Robot Learning}.\hskip 1em plus 0.5em minus 0.4em\relax London, UK: PMLR, 8--11 November 2021, pp. 432--448.

\bibitem{lee2021learning}
R.~Lee, D.~Ward, V.~Dasagi, A.~Cosgun, J.~Leitner, and P.~Corke, ``Learning arbitrary-goal fabric folding with one hour of real robot experience,'' in \emph{Conference on Robot Learning}.\hskip 1em plus 0.5em minus 0.4em\relax London, UK: PMLR, 11 November 2021, pp. 2317--2327.

\bibitem{moerland2020framework}
T.~M. Moerland, J.~Broekens, and C.~M. Jonker, ``A framework for reinforcement learning and planning,'' \emph{arXiv preprint arXiv:2006.15009}, 2020.

\bibitem{ebert2018visual}
\BIBentryALTinterwordspacing
F.~Ebert, C.~Finn, S.~Dasari, A.~Xie, A.~X. Lee, and S.~Levine, ``Visual foresight: Model-based deep reinforcement learning for vision-based robotic control,'' \emph{CoRR}, vol. abs/1812.00568, 2018. [Online]. Available: \url{http://arxiv.org/abs/1812.00568}
\BIBentrySTDinterwordspacing

\bibitem{pfaff2021learning}
\BIBentryALTinterwordspacing
T.~Pfaff, M.~Fortunato, A.~Sanchez-Gonzalez, and P.~Battaglia, ``Learning mesh-based simulation with graph networks,'' in \emph{International Conference on Learning Representations}, 2021. [Online]. Available: \url{https://openreview.net/forum?id=roNqYL0_XP}
\BIBentrySTDinterwordspacing

\bibitem{huang2023self}
Z.~Huang, X.~Lin, and D.~Held, ``Self-supervised cloth reconstruction via action-conditioned cloth tracking,'' \emph{arXiv preprint arXiv:2302.09502}, 2023.

\bibitem{canberk2023cloth}
A.~Canberk, C.~Chi, H.~Ha, B.~Burchfiel, E.~Cousineau, S.~Feng, and S.~Song, ``Cloth funnels: Canonicalized-alignment for multi-purpose garment manipulation,'' in \emph{2023 IEEE International Conference on Robotics and Automation (ICRA)}.\hskip 1em plus 0.5em minus 0.4em\relax IEEE, 2023, pp. 5872--5879.

\bibitem{wang2023trtm}
W.~Wang, G.~Li, M.~Zamora, and S.~Coros, ``Trtm: Template-based reconstruction and target-oriented manipulation of crumpled cloths,'' \emph{arXiv preprint arXiv:2308.04670}, 2023.

\bibitem{lee2020stochastic}
A.~X. Lee, A.~Nagabandi, P.~Abbeel, and S.~Levine, ``Stochastic latent actor-critic: Deep reinforcement learning with a latent variable model,'' \emph{Advances in Neural Information Processing Systems}, vol.~33, pp. 741--752, 2020.

\bibitem{haarnoja2018soft}
T.~Haarnoja, A.~Zhou, P.~Abbeel, and S.~Levine, ``Soft actor-critic: Off-policy maximum entropy deep reinforcement learning with a stochastic actor,'' in \emph{International conference on machine learning}.\hskip 1em plus 0.5em minus 0.4em\relax Stockholm, Sweden: PMLR, 10--15 July 2018, pp. 1861--1870.

\bibitem{ziebart2010modeling}
B.~D. Ziebart, J.~A. Bagnell, and A.~K. Dey, ``Modeling interaction via the principle of maximum causal entropy,'' in \emph{ICML}, 2010.

\bibitem{levine2018reinforcement}
S.~Levine, ``Reinforcement learning and control as probabilistic inference: Tutorial and review,'' \emph{arXiv preprint arXiv:1805.00909}, 2018.

\bibitem{kaelbling1998planning}
L.~P. Kaelbling, M.~L. Littman, and A.~R. Cassandra, ``Planning and acting in partially observable stochastic domains,'' \emph{Artificial intelligence}, vol. 101, no. 1-2, pp. 99--134, 1998.

\bibitem{lee2018stochastic}
A.~X. Lee, R.~Zhang, F.~Ebert, P.~Abbeel, C.~Finn, and S.~Levine, ``Stochastic adversarial video prediction,'' \emph{arXiv preprint arXiv:1804.01523}, 2018.

\bibitem{weng2022fabricflownet}
T.~Weng, S.~M. Bajracharya, Y.~Wang, K.~Agrawal, and D.~Held, ``Fabricflownet: Bimanual cloth manipulation with a flow-based policy,'' in \emph{Conference on Robot Learning}.\hskip 1em plus 0.5em minus 0.4em\relax uckland, New Zealand: PMLR, 15–18 December 2022, pp. 192--202.

\bibitem{laskin2020curl}
M.~Laskin, A.~Srinivas, and P.~Abbeel, ``Curl: Contrastive unsupervised representations for reinforcement learning,'' in \emph{International Conference on Machine Learning}.\hskip 1em plus 0.5em minus 0.4em\relax PMLR, 2020, pp. 5639--5650.

\bibitem{kostrikov2020image}
I.~Kostrikov, D.~Yarats, and R.~Fergus, ``Image augmentation is all you need: Regularizing deep reinforcement learning from pixels,'' \emph{arXiv preprint arXiv:2004.13649}, 2020.

\end{thebibliography}

\end{document}